\documentclass[sigconf]{acmart}

\AtBeginDocument{%
  }
\usepackage{color}

\usepackage{soul}
\usepackage{subcaption}

\setcopyright{cc}
\setcctype{by}
\acmDOI{10.1145/3776734.3794444}
\acmYear{2026}
\copyrightyear{2026}
\acmISBN{979-8-4007-2321-6/2026/03}
\acmConference[HRI Companion '26]{Companion Proceedings of the 21st ACM/IEEE International Conference on Human-Robot Interaction}{March 16--19, 2026}{Edinburgh, Scotland, UK}
\acmBooktitle{Companion Proceedings of the 21st ACM/IEEE International Conference on Human-Robot Interaction (HRI Companion '26), March 16--19, 2026, Edinburgh, Scotland, UK}

\begin{document}

\title[A User Study on the Suitability of Teleoperation Interfaces for Primitive Manipulation Tasks]{A User Study on the Suitability of Teleoperation Interfaces \\for Primitive Manipulation Tasks}

\author{Jun Aoki}
\authornotemark[1]
\email{j.aoki@aist.go.jp}
\affiliation{%
  \institution{AIST}
  \city{Koto}
  \state{Tokyo}
  \country{Japan}
}
\affiliation{%
  \institution{University of Tsukuba}
  \city{Tsukuba}
  \state{Ibaraki}
  \country{Japan}
}

\author{Shunki Itadera}
\affiliation{%
  \institution{National Institute of Advanced Industrial Science and Technology (AIST)}
  \city{Koto}
  \state{Tokyo}
  \country{Japan}
}

\renewcommand{\shortauthors}{Aoki and Itadera}%Trovato et al.}

\begin{abstract}
  The application of teleoperation to control robotic arms has been widely explored, and user-friendly teleoperation systems have been studied for facilitating higher performance and lower operational burden. To investigate the dominant factors in a practical teleoperation system, this study focused on the characteristics of an interface used to operate a robotic arm.
  The usability of an interface depends on the characteristics of the manipulation tasks to be completed; however, systematic comparisons of different interfaces across different tasks remain limited. In this study, we compared two widely used teleoperation interfaces, a 3D mouse and a VR controller, for two simple yet broadly applicable tasks with a six-degree-of-freedom (6DoF) robotic arm: repetitively pushing buttons and rotating knobs. Participants (N = 23) controlled a robotic arm with 6DoF to push buttons and rotate knobs as many times as possible in 3-minute trials. Each trial was followed by a NASA\mbox{-}TLX workload rating. The results showed a clear connection between the interface and task performance: the VR controller yielded higher performance for pushing buttons, whereas the 3D mouse performed better and was less demanding for knob rotation. These findings highlight the importance of considering dominant motion primitives of the task when designing practical teleoperation interfaces.
\end{abstract}


\begin{teaserfigure}
  \centering
  \begin{minipage}[b]{0.6\columnwidth}
    \centering
    \includegraphics[width=0.86\columnwidth,alt={Diagram showing the remote control setup. This diagram illustrates the system overview of the two primary interfaces used in the research.}]{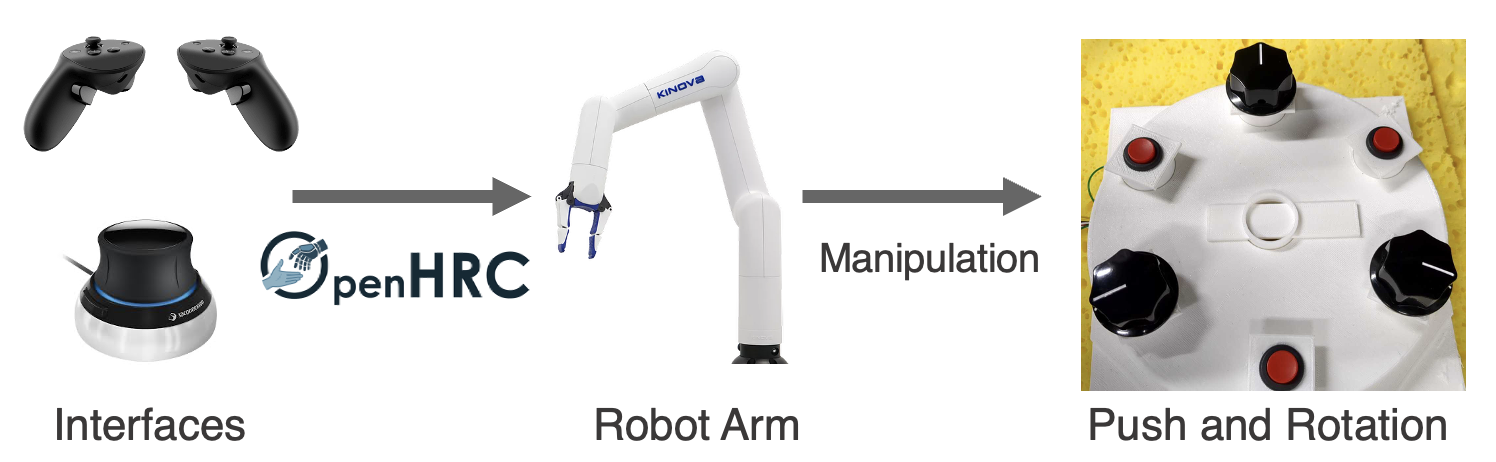}
    \caption*{(a) System overview}
  \end{minipage}
  \begin{minipage}[b]{0.3\columnwidth}
    \centering
    \includegraphics[width=0.88\columnwidth,alt={Diagram showing the remote control setup. Participant operates the robot using a VR controller.}]{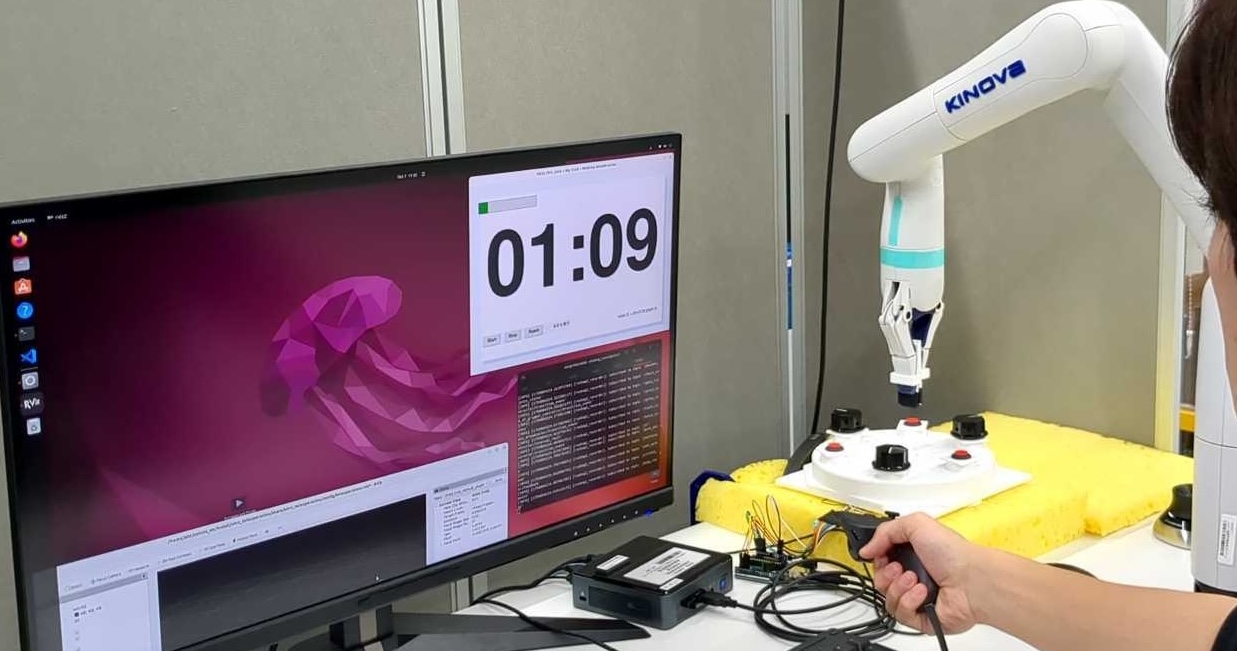}
    \caption*{(b) Experiment with the VR controller}
  \end{minipage}
  \caption{(a) System overview. (b) Experimental setup with the VR controller}
  \label{fig:system_overflow}
\end{teaserfigure}

\maketitle
\section{Introduction}

Teleoperation is increasingly used for assistive manipulation, remote inspection, and data collection for robot learning from demonstration~\cite{Luo2020,Mandlekar2018,li2025modality}, making interface design critical because the choice of interface affects not only task performance but also the quality and quantity of collected demonstrations. Designing interfaces that are efficient, intuitive, and not overly fatiguing is therefore a key challenge.

Several classes of devices are used in teleoperation
interfaces for robotic arms. For example, a 3D mouse providing input with six-degree-of-freedom (6DoF) via a compact desktop device has been used for precise
control of manipulator devices in web-based teleoperation~\cite{Dhat2024Using3DMouse}.
VR controllers track the pose of the operator’s hand and are widely
used in VR-based teleoperation systems for mobile and aerial
manipulators~\cite{sobhani2020vamhri,Meng2023,Yang2025VRUAVTeleop}.
GUI-based interfaces relying on a 2D display with keyboard-and-mouse
or gamepad input remain common, and have been directly compared against VR in robot grasping and mixed-reality digital-twin control~\cite{lemansurier2024roman,Wu2025DigitalTwinTeleop}. Beyond these mainstream devices, more novel interfaces have also been explored, including markerless hand/arm tracking, brain–computer interfaces, and sketch-based commands~\cite{kofmanMarkerless,Xie2024MediaPipeTeleop,EEG,Arshad2022BCIArm,sakamotoSketch}, typically in narrowly defined tasks with specialized setups. Overall, prior works have tended to evaluate a single interface on a narrow family of tasks. Thus, systematic evidence on how interface characteristics interact with different types of tasks remains limited.

In this study, we take a step toward such a comparison by focusing on two basic yet widely applicable primitive manipulation tasks: pushing a button and rotating a knob. On a standardized circular switch platform (Fig.~\ref{fig:system_overflow}a), participants controlled a 6DoF robotic arm using either a compact 3D mouse, which provides single-device 6DoF rate control~\cite{Dhat2024Using3DMouse}, or a pose-tracked VR motion controller for gesture-based, hand-centered control ~\cite{sobhani2020vamhri,Meng2023}.
We measured the number of successful switch operations within a fixed time window and the subjective workload using NASA\mbox{-}TLX. After practice and an initial familiarization session that also included a GUI interface, we focused our analysis on a second session in which participants reused the 3D mouse and VR controller after becoming familiar with the tasks. Our results revealed a clear connection between the task and the interface, in that the VR controller was more effective for pushing buttons, whereas the 3D mouse was more effective for rotating a knob.

\section{Method}

\subsection{Apparatus}

We used a robotic arm with 6DoF (Kinova Gen3 Lite; Kinova Robotics) equipped with a gripper. Two main teleoperation interfaces were used to control the pose of the robot's gripper, specifically a 3D mouse (3Dconnexion) and a VR controller (Meta Quest Pro). The 3D mouse interface also allowed the users to provide input with 6DoF, and the VR controller interface tracked the pose of the participant's dominant hand.

All interfaces were implemented on top of the open-source framework OpenHRC~\cite{OpenHRC}, which converted device inputs into Cartesian end-effector velocity commands for the robot. The opening and closing motions of the gripper were controlled via keyboard keys, identically across interfaces. For both the 3D mouse and VR controller, three translational and three rotational velocity components were mapped to the same end-effector frame, and gains and step sizes were tuned so that similar inputs produced comparable translational and rotational speeds across interfaces.

\subsection{Switch platform}

The task platform consisted of push-buttons and rotary knobs arranged on a circular 3D-printed base placed on a table (Fig.~\ref{fig:system_overflow}b). The buttons were small tactile switches with a short stroke; the knobs could be rotated by at least 270°. The buttons and knobs were placed at equal angular intervals along a circle.

A switch operation was considered successful when the button was fully depressed (pushing task) or the knob had been rotated by 180° (rotation task). In both cases, a buzzer was activated on success to provide immediate feedback.

\subsection{Participants}

We recruited 23 participants (age: 20-65 years; 10 were female) from outside of our institute, with mixed prior experience in robot operation. All participants used their dominant hand and reported normal or corrected to normal vision. Written informed consent was obtained prior to participation. This experiment was approved by the ethics committee of AIST (2024-1434A).

Of the 23 participants, one participant was unable to complete the rotation task due to scheduling issues and was excluded from analysis of the rotation task (described in Section~\ref{sec:tasks}). Two participants were observed to rub the side of the knob without producing 180° rotations, leading to overestimated success scores. Because this behavior violated the task instructions, these two participants were treated as protocol violations and were excluded from the rotation task analysis ($N = 20$). Including these two participants did not change the qualitative pattern of results.

\subsection{Tasks}
\label{sec:tasks}
We defined two tasks:

\begin{itemize}
  \item \textbf{Pushing task}: Participants were instructed to push buttons arranged in a circle, moving in a clockwise order, and to complete as many successful pushes as possible within 3 minutes.
  \item \textbf{Rotation task}: Participants were instructed to rotate knobs arranged in a circle, again in a clockwise order, as many times as possible within 3 minutes. A success was counted when a knob had been rotated by 180°. Participants were instructed not to keep rotating knobs in the same direction indefinitely but to alternate the rotation direction relative to the previous knob to avoid joint limits.
\end{itemize}

\subsection{Design and procedure}
\label{sec:design}

The experiment comprised two sessions. In Session~1, participants completed practice trials with all three interfaces (GUI, 3D mouse, VR controller) after receiving task instructions. This session was designed to familiarize participants with the pushing and rotation tasks and with the three interfaces.

In Session~2, participants performed the pushing and rotation tasks again using only the 3D mouse and VR controller. The order of the two interfaces was counterbalanced across participants. All performance and workload results reported in the following sections are based on Session~2, after participants had become familiar with the tasks and interfaces.

After each trial, participants completed the NASA\mbox{-}TLX questionnaire to rate their subjective workload.

\begin{figure*}[t]
  \centering
  \begin{minipage}{.48 \linewidth}
    \centering
    \includegraphics[width= \linewidth,alt={Box-and-whisker plot comparing 3D mouse and VR controller performance in button presses over 3 minutes.}]{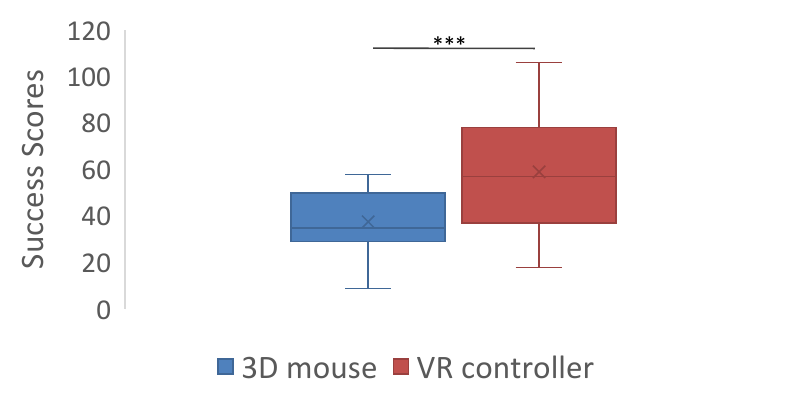}
    \subcaption{Success Scores in Pushing Task ($N = 23$)}
    \label{fig:push_success}
  \end{minipage}%
  \begin{minipage}{.48 \linewidth}
    \centering
    \includegraphics[width= \linewidth,alt={Box-and-whisker plot comparing 3D mouse and VR controller performance in rotations over 3 minutes.}]{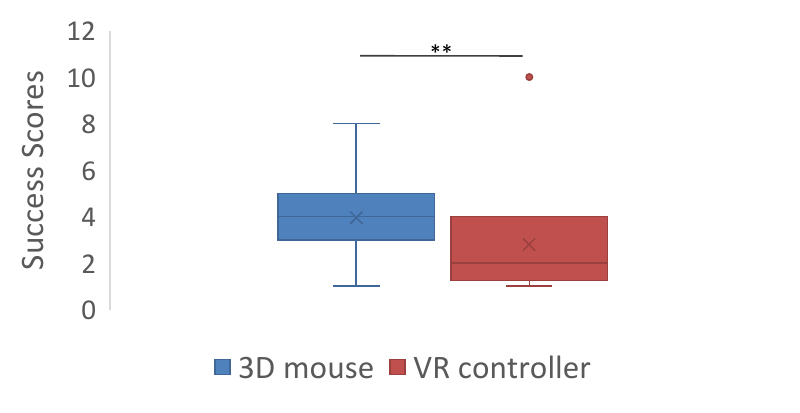}
    \subcaption{Success Scores in Rotation Task ($N = 20$)}
    \label{fig:rotation_success}
  \end{minipage}
  \caption{Success scores in the pushing and rotation tasks. Error bars show $\pm 1$ SE. ** $p < .01$, *** $p < .001$}
  \label{fig:performance}
\end{figure*}

\begin{figure*}[t]
  \centering
  \begin{minipage}{.48 \linewidth}
    \centering
    \includegraphics[width= \linewidth,alt={Box-and-whisker plot of weighted overall NASA-TLX workload scores for push operations comparing 3D mouse and VR controllers.}]{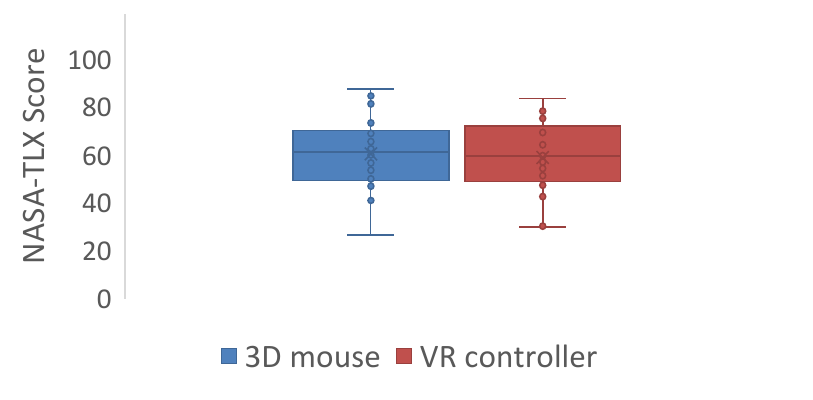}
    \subcaption{NASA\mbox{-}TLX Scores in Pushing Task ($N = 23$)}
    \label{fig:push_tlx}
  \end{minipage}%
  \begin{minipage}{.48 \linewidth}
    \centering
    \includegraphics[width= \linewidth,alt={Box-and-whisker plot of weighted overall NASA-TLX workload scores for rotation operations comparing 3D mouse and VR controllers.}]{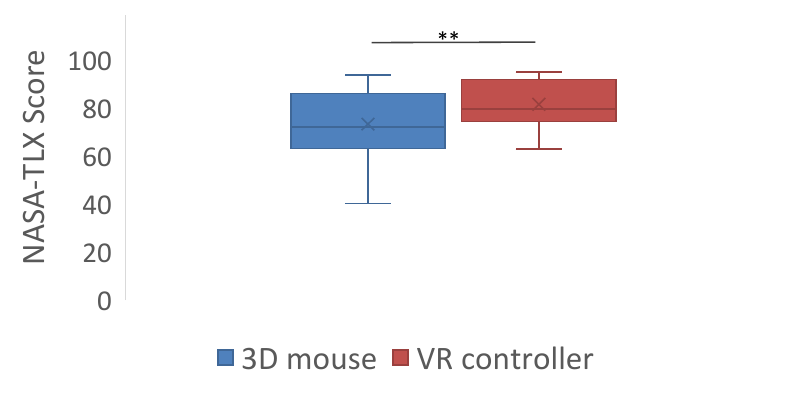}
    \subcaption{NASA\mbox{-}TLX Scores in Rotation Task ($N = 20$)}
    \label{fig:rotation_tlx}
  \end{minipage}
  \caption{Weighted overall NASA\mbox{-}TLX scores in the pushing and rotation tasks. Error bars show $\pm 1$ SE. ** $p < .01$}
  \label{fig:workload}
\end{figure*}

\subsection{Measures and analysis}

To evaluate the participants' subjective workload, we computed weighted overall NASA\mbox{-}TLX scores for each condition based on pairwise comparisons, and also inspected individual subscales to identify dimensions where interfaces differed. For comparisons between the 3D mouse and VR controller, we used two-tailed paired $t$-tests in Session~2. Subscale analyses were exploratory and uncorrected for multiple comparisons.

\section{Results}

\subsection{Performance}
We report performance results only from Session~2. Fig.~\ref{fig:performance} shows the number of successful operations completed within the 3-minute time limit for
the pushing and rotation tasks.

For the pushing task, participants completed more successful
pushes with the VR controller than with the 3D mouse (paired
$t(22) = -4.58$, $p < .001$, Cohen's $d = -0.96$). For the
rotation task, the 3D mouse yielded more successful 180° knob
rotations than the VR controller (excluding two protocol
violations), and paired tests supported the superiority of the
3D mouse ($t(19) = 3.24$, $p = .006$, Cohen's $d = 0.69$).

\subsection{Subjective workload}

Fig.~\ref{fig:workload} shows NASA\mbox{-}TLX overall scores for the
3D mouse and VR controller in Session~2. On the pushing task,
there was no statistically significant difference in overall workload
between the VR controller and 3D mouse ($t(22) = 0.82$, $p = .42$,
Cohen's $d = 0.17$). In contrast, for the rotation task, the 3D mouse showed significantly lower workload than the VR controller
($t(19) = -3.01$, $p = .007$, Cohen's $d = -0.67$), consistent with the performance advantage observed in success scores.

\begin{figure*}[t]
  \begin{minipage}[b]{0.49\textwidth}
    \centering
    \includegraphics[width=\linewidth,alt={Group-by-group average bar graph of NASA–TLX subscales. Compares 3D mouse and VR controller usage, showing results for pushing task.}]{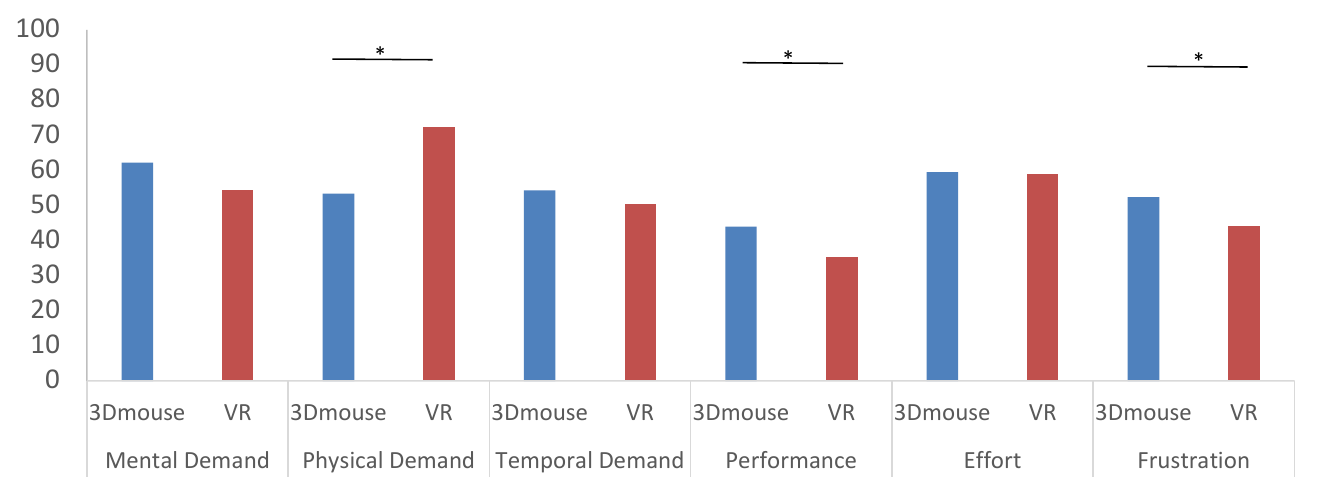}
    \caption*{(a) Pushing task ($N = 23$)}
  \end{minipage}
  \hfill
  \begin{minipage}[b]{0.49\textwidth}
    \centering
    \includegraphics[width=\linewidth,alt={Group-by-group average bar graph of NASA–TLX subscales. Compares 3D mouse and VR controller usage, showing results for rotation task.}]{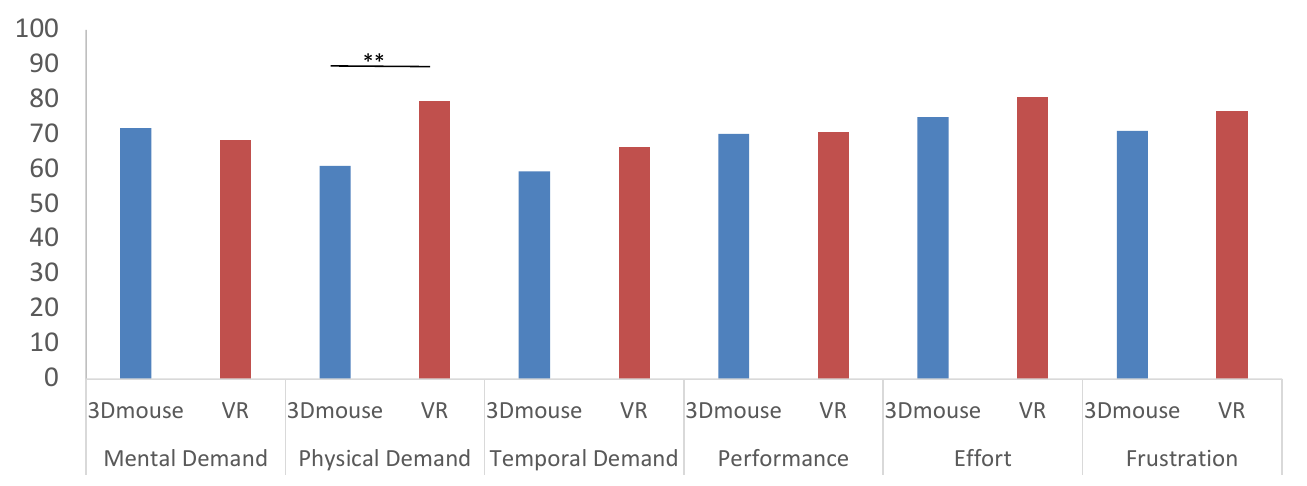}
    \caption*{(b) Rotation task ($N = 20$)}
  \end{minipage}
  \vspace{-2mm}
  \caption{NASA\mbox{-}TLX subscale scores in the pushing and rotation tasks. Error bars show $\pm 1$ SE. * $p < .05$, ** $p < .01$}
  \label{fig:tlx_subscales}
\end{figure*}

As shown in Fig.~\ref{fig:tlx_subscales}, the average scores on the NASA\mbox{-}TLX subscales clarify these patterns. Across both tasks, physical demand ratings were higher for the VR controller than for the 3D mouse (both $p < .05$), which indicates that the pose-tracking control imposed a greater physical demand. In the rotation task, this higher physical demand for the VR controller was reflected in the overall NASA\mbox{-}TLX scores. In the pushing task, however, the 3D mouse was rated as more frustrating and yielded a poorer self-rated performance than the VR controller ($p < .05$). These opposing tendencies showed that the higher physical demand in using the VR controller and the higher frustration with and worse perceived performance of the 3D mouse partially offset each other in the weighted NASA\mbox{-}TLX scores, resulting in no significant differences in overall workload for the pushing task.

\section{Discussion}

We compared a 3D mouse and a VR controller as teleoperation
interfaces for a robotic arm with 6DoF on two simple yet widely
applicable tasks, including repeatedly pushing buttons (primarily translational) and repeatedly rotating knobs (primarily rotational).

Our main finding was a clear contrast in which interface was more effective for each primitive manipulation task.
The VR controller was more effective for pushing buttons, whereas
the 3D mouse was more effective in rotating the knobs. This pattern aligns with recent calls for task-aware evaluations of teleoperation interfaces, which emphasize that interface effectiveness should be understood in the context of specific tasks and operator capabilities rather than in isolation ~\cite{Rea2022UserCenteredTeleop}.

\subsection{Suitability for primitive manipulation tasks}

In the button-pushing task, the participants achieved higher success scores with the VR controller compared to the 3D mouse. Mapping the controller pose directly to the end-effector motion allowed natural reaching and wrist movements for fast button alignment and pushing. Similar benefits of intuitive pose-based control have been reported in
VR-based telerobotics, in which more natural interaction designs
improved task performance and reduced subjective
workload~\cite{Nenna2023}.

In contrast, the rotation task favored the 3D mouse. Generating repeated 180° knob rotations with the VR controller required large or segmented wrist motions, which were physically demanding and
made it easy for the participants to drift away from the ideal rotation axis. With the 3D mouse, participants performed stable continuous rotations using small, sustained deflections of the device, which led to more
successful rotations. Although these rotation-task differences were smaller than the pushing-task contrast, they suggest that motion primitives requiring sustained rotation may benefit from interfaces that decouple rotation from arm or wrist movements\cite{Dhat2024Using3DMouse}.

From a design perspective, our results highlight the importance of
matching teleoperation interfaces to the dominant motion
primitives of the task. For tasks that are dominated by translational reaching and discrete contacts (e.g., pushing buttons, picking up and placing objects), VR controllers or other pose-tracked hand interfaces appear
promising after minimal practice, in line with prior VR-based teleoperation work on mobile and aerial manipulators
\cite{sobhani2020vamhri,Meng2023,Yang2025VRUAVTeleop}.
Related studies on pick-and-place and legged manipulator
teleoperation further suggest that immersive or mixed-reality interfaces can improve spatial understanding and efficiency when the task is dominated by reaching and placement \cite{Kumar2025,Ulloa2024}.
For tasks that require sustained rotational
motion, 3D mouse style interfaces may provide more stable control with less physical fatigue. More broadly, our results illustrate how teleoperation tasks should guide interface selection and adaptation in practical
deployments~\cite{Rea2022UserCenteredTeleop}.

\subsection{Workload profiles and practical implications}

Subjective workload patterns generally aligned with interface
suitability and revealed important trade-offs between subscales. In
the rotation task, the 3D mouse yielded significantly lower overall
NASA\mbox{-}TLX scores than the VR controller, consistent with its higher
success scores. The subscale analysis showed that the VR controller
was rated as imposing substantially higher physical demand in both tasks, particularly for rotation, which suggests that repeated wrist
rotations incurred physical fatigue that likely contributed to the
higher overall workload. Comparable modality-dependent workload differences have been reported in industrial robot teleinspection, grasping and digital-twin teleoperation~\cite{Mazeas2025,lemansurier2024roman,Wu2025DigitalTwinTeleop}.

In the pushing task, by contrast, overall NASA\mbox{-}TLX scores did not differ significantly between interfaces, although the VR
controller achieved higher success scores. The subscale scores indicate that this did not occur because the VR controller was less
demanding in general, but because different workload dimensions pulled in opposite directions: physical demand was higher for the VR controller, whereas frustration and performance indicators were
worse for the 3D mouse. That is, users perceived the VR
controller as physically effortful but effective, and the 3D mouse as less physically demanding but more frustrating and harder to use successfully. Considering only the overall NASA\mbox{-}TLX score would therefore obscure these trade-offs.

More broadly, our workload patterns are consistent with prior reviews of mixed-reality teleoperation, which report persistent physical demand and frustration in challenging tasks and show that performance metrics and subjective workload can capture such differences more reliably than physiological indicators~\cite{Stacy2021Workload,Odoh2024}.
Teleoperation interface design should therefore explicitly consider
multi-dimensional workloads rather than only success rates or completion time, and should evaluate whether physical demand, frustration, or perceived performance is the dominant limiting factor for a given task.

\subsection{Limitations and future work}

This study has several limitations. First, we focused on simple switch-based tasks with a fixed 3-minute time limit; while this isolates basic motion primitives, real teleoperation often involves longer, multi-step activities with more complex objects. Second, we evaluated performance mainly via success scores and subjective workload, without analyzing detailed motion characteristics or fully treating the GUI interface as a baseline condition.

Future work should extend the evaluation to more realistic compound tasks, incorporate richer objective metrics, and examine learning over longer periods and across a broader set of interfaces. Our results also motivate hybrid or adaptive interfaces that combine or switch modalities to reduce operator workload ~\cite{Mazeas2025,Kumar2025}.

\begin{acks}
  This work was supported in part by JST CRONOS JPMJCS24K6 and BRIDGE/Practical Global Research in the AI x Robotics Services by CAO Japan.
\end{acks}

\bibliographystyle{ACM-Reference-Format}
\bibliography{reference}

@article{Luo2020,
  author  = {Luo, J. and others},
  title   = {Combined perception, control, and learning for teleoperation: key technologies, applications, and challenges},
  journal = {Cognitive Computation and Systems},
  volume  = {2},
  number  = {2},
  pages   = {33--43},
  year    = {2020}
}

@inproceedings{Mandlekar2018,
  author    = {Mandlekar, A. and others},
  title     = {{RoboTurk}: A crowdsourcing platform for robotic skill learning through imitation},
  booktitle = {Proceedings of the 2nd Conference on Robot Learning (CoRL)},
  series    = {Proceedings of Machine Learning Research (PMLR)},
  volume    = {87},
  pages     = {879--893},
  year      = {2018}
}

@misc{li2025modality,
  author        = {Li, H. and Cui, Y. and Sadigh, D.},
  title         = {How to train your robots? The impact of demonstration modality on imitation learning},
  year          = {2025},
  archiveprefix = {arXiv},
  eprint        = {2503.07017}
}

@inproceedings{Dhat2024Using3DMouse,
  author    = {Dhat, V. and Walker, N. and Cakmak, M.},
  title     = {Using {3D} mice to control robot manipulators},
  booktitle = {Proceedings of the ACM/IEEE International Conference on Human--Robot Interaction (HRI)},
  address   = {Boulder, CO, USA},
  pages     = {1--5},
  year      = {2024}
}

@article{Yang2025VRUAVTeleop,
  author  = {Yang, Z. and Tomita, K. and Kamimura, A.},
  title   = {{VR}-based teleoperation of {UAV}--manipulator systems: From single-{UAV} control to dual-{UAV} cooperative manipulation},
  journal = {Applied Sciences},
  volume  = {15},
  number  = {20},
  pages   = {11086},
  year    = {2025}
}

@inproceedings{sobhani2020vamhri,
  author    = {Sobhani, M. and others},
  title     = {Robot teleoperation through virtual reality interfaces},
  booktitle = {Proceedings of the {VAM-HRI} Workshop},
  year      = {2020}
}

@article{Meng2023,
  author  = {Meng, L. and others},
  title   = {{VR}-based robot teleoperation via human-centered interfaces},
  journal = {Procedia Computer Science},
  volume  = {217},
  pages   = {130--139},
  year    = {2023}
}

@article{Wu2025DigitalTwinTeleop,
  author  = {Wu, Y. and Zhao, B. and Li, Q.},
  title   = {The teleoperation of robot arms by interacting with an object’s digital twin in a mixed reality environment},
  journal = {Applied Sciences},
  volume  = {15},
  number  = {7},
  pages   = {3549},
  year    = {2025}
}

@inproceedings{lemansurier2024roman,
  author    = {LeMasurier, G. and others},
  title     = {Comparing {2D} keyboard-and-mouse to virtual reality for robot grasping},
  booktitle = {Proceedings of the {IEEE} International Conference on Robot and Human Interactive Communication ({RO-MAN})},
  year      = {2024}
}

@inproceedings{Xie2024MediaPipeTeleop,
  author    = {Xie, J. and Xu, Z. and Zeng, J. and Du, X. and Zhang, Y. and Wang, S. and Chen, H. and Hashimoto, K.},
  title     = {A novel teleoperation approach based on {MediaPipe} and {LSTM}},
  booktitle = {Proceedings of the 7th {JC-IFToMM} International Symposium on Theory of Machines and Mechanisms},
  address   = {Kitakyushu, Japan},
  pages     = {178--185},
  year      = {2024}
}

@article{kofmanMarkerless,
  author  = {Kofman, J. and Verma, S. and Wu, X.},
  title   = {Robot-manipulator teleoperation by markerless vision-based hand--arm tracking},
  journal = {International Journal of Optomechatronics},
  volume  = {1},
  number  = {3},
  pages   = {331--357},
  year    = {2007}
}

@article{Arshad2022BCIArm,
  author  = {Arshad, J. and Qaisar, A. and Rehman, A.-U. and Shakir, M. and Nazir, M. K. and Rehman, A. U. and Eldin, E. T. and Ghamry, N. A. and Hamam, H.},
  title   = {Intelligent control of robotic arm using brain computer interface and artificial intelligence},
  journal = {Applied Sciences},
  volume  = {12},
  number  = {21},
  pages   = {10813},
  year    = {2022}
}

@article{EEG,
  author  = {Xu, B. and Li, W. and He, X. and Wei, Z. and Zhang, D. and Wu, C. and Song, A.},
  title   = {Motor imagery based continuous teleoperation robot control with tactile feedback},
  journal = {Electronics},
  volume  = {9},
  number  = {1},
  pages   = {174},
  year    = {2020}
}

@inproceedings{sakamotoSketch,
  author    = {Sakamoto, D. and Honda, K. and Inami, M. and Igarashi, T.},
  title     = {Sketch and run: A stroke-based interface for home robot},
  booktitle = {Proceedings of the 27th International Conference on Human Factors in Computing Systems (CHI)},
  pages     = {197--200},
  year      = {2009}
}

@misc{OpenHRC,
  author = {Itadera, S.},
  title  = {Cross-robot-interface teleoperation framework towards implementation-friendly human--robot collaboration},
  note   = {TechRxiv preprint},
  year   = {2025}
}

@article{Rea2022UserCenteredTeleop,
  author  = {Rea, D. J. and Seo, S. H.},
  title   = {Still not solved: A call for renewed focus on user-centered teleoperation interfaces},
  journal = {Frontiers in Robotics and {AI}},
  volume  = {9},
  pages   = {704225},
  year    = {2022}
}

@article{Nenna2023,
  author  = {Nenna, F. and Zanardi, D. and Gamberini, L.},
  title   = {Enhanced interactivity in {VR}-based telerobotics: An eye-tracking investigation of human performance and workload},
  journal = {International Journal of Human--Computer Studies},
  volume  = {177},
  pages   = {103079},
  year    = {2023}
}

@misc{Kumar2025,
  author        = {Kumar, A. and Simangunsong, S. and Carreno-Medrano, P. and Cosgun, A.},
  title         = {Mixed reality outperforms virtual reality for remote error resolution in pick-and-place tasks},
  year          = {2025},
  archiveprefix = {arXiv},
  eprint        = {2502.06141}
}

@article{Ulloa2024,
  author  = {Cruz Ulloa, C. C. and Dominguez, D. and del Cerro, J. and Barrientos, A.},
  title   = {Analysis of {MR}--{VR} tele-operation methods for legged--manipulator robots},
  journal = {Virtual Reality},
  volume  = {28},
  pages   = {131},
  year    = {2024}
}

@article{Mazeas2025,
  author  = {Mazeas, D. and Namoano, B.},
  title   = {Study of visualization modalities on industrial robot teleoperation for inspection in a virtual co-existence space},
  journal = {Virtual Worlds},
  volume  = {4},
  number  = {2},
  pages   = {17},
  year    = {2025}
}

@article{Stacy2021Workload,
  author  = {Stacy, A. and Hancock, M.},
  title   = {Workload in mixed reality teleoperation: A review},
  journal = {{IEEE} Transactions on Visualization and Computer Graphics},
  volume  = {27},
  number  = {5},
  pages   = {1--13},
  year    = {2021}
}

@article{Odoh2024,
  author  = {Odoh, G. and Landowska, A. and Crowe, E. M. and Benali, K. and Cobb, S. and Wilson, M. L. and Maior, H. A. and Kucukyilmaz, A.},
  title   = {Performance metrics outperform physiological indicators in robotic teleoperation workload assessment},
  journal = {Scientific Reports},
  volume  = {14},
  pages   = {30984},
  year    = {2024}
}

\end{document}